\documentclass[letterpaper, 10 pt, conference]{ieeeconf}

\IEEEoverridecommandlockouts                             

\usepackage{amsmath,amsfonts}
\usepackage{array}
\usepackage{float}
\usepackage[caption=false,font=normalsize, labelfont=sf,textfont=sf]{subfig}
\usepackage{hyperref}
\usepackage{textcomp}
\usepackage{graphicx}
\usepackage{stfloats}
\usepackage{cite}
\usepackage{threeparttable}
\usepackage{booktabs}
\usepackage{color}
\usepackage{url}

\begin{document}


\title{ Streamlining Forest Wildfire Surveillance: AI-Enhanced UAVs Utilizing the FLAME Aerial Video Dataset for Lightweight and Efficient Monitoring}

\author{\authorblockN{\thanks{This work was jointly supported by JSPS KAKENHI (Grants-in-Aid for Scientific Research, 21H05001); the Major Project of the MOE (China) National Key Research Bases for Humanities and Social Sciences (22JJD910003); Beijing Golden Bridge Project seed fund. Project No.ZZ21021; This work was jointly supported by the Co-creation Center for Disaster Resilience, Tohoku University, and the Public Computing Cloud, Renmin University of China.}\thanks{Yanbing Bai is an assistant professor at the Center for Applied Statistics and School of Statistics, Renmin University of China, Beijing, 100872, China, e-mail: ybbai@ruc.edu.cn. Corresponding author} \thanks{Lemeng Zhao is a student at the Center for Applied Statistics and School of Statistics, Renmin University of China, Beijing, 100872, China, e-mail: zhaolemeng@ruc.edu.cn} \thanks{Junjie Hu is a researcher at Shenzhen Institute of Artificial Intelligence and Robotics for Society, Shenzhen, 518129, China, e-mail: hujunjie@cuhk.edu.cn} \thanks{Jianchao Bi is a student at Gaoling School of Artificial Intelligence, Renmin University of China, Beijing, 100872, China, e-mail: bijianchao@ruc.edu.cn} \thanks{Erick Mas is an Associate Professor with the International Research Institute of Disaster Science (IRIDeS), Tohoku University, Sendai, 980-8572, Japan, e-mail: mas@tohoku.ac.jp} \thanks{Shunichi Koshimura is a Professor with the International Research Institute of Disaster Science (IRIDeS), Tohoku University, Sendai, 980-8572, Japan, e-mail: koshimura@tohoku.ac.jp. Corresponding author} Lemeng Zhao, Junjie Hu, Jianchao Bi, Yanbing Bai\textsuperscript{*},~\IEEEmembership{Member,~IEEE,}\\ Erick Mas, Shunichi Koshimura\textsuperscript{*}}}

\maketitle
\thispagestyle{empty}
\pagestyle{empty}

\begin{abstract}
In recent years, unmanned aerial vehicles (UAVs) have played an increasingly crucial role in supporting disaster emergency response efforts by analyzing aerial images. While current deep-learning models focus on improving accuracy, they often overlook the limited computing resources of UAVs. This study recognizes the imperative for real-time data processing in disaster response scenarios and introduces a lightweight and efficient approach for aerial video understanding. Our methodology identifies redundant portions within the video through policy networks and eliminates this excess information using frame compression techniques. Additionally, we introduced the concept of a `station point,' which leverages future information in the sequential policy network, thereby enhancing accuracy. To validate our method, we employed the wildfire FLAME dataset. Compared to the baseline, our approach reduces computation costs by more than 13 times while boosting accuracy by 3$\%$. Moreover, our method can intelligently select salient frames from the video, refining the dataset. This feature enables sophisticated models to be effectively trained on a smaller dataset, significantly reducing the time spent during the training process.
\end{abstract}

\IEEEpeerreviewmaketitle

\section{Introduction}
Forest fires, posing a significant threat to human life and the ecological environment, have garnered increased attention in recent years \cite{bouguettaya2022review}. In particular, UAV-based platforms are increasingly used for post-disaster monitoring and response due to their ability to acquire real-time and efficient ground information\cite{ozkan2021optimization, nagasawa2021model}. This approach enhances data acquisition speed and reduces the risks associated with manual inspection of forest fires, ensuring the safety of lives.

Recently, deep learning algorithms have been at the forefront of detecting forest fires using visual cues like flame \cite{goyal2020yolo} or smoke \cite{alexandrov2019analysis, hossain2020forest,zhan2021pdam}, drawing extensive attention within the research community, in particular, models for video stream data understanding\cite{bouguettaya2022review}. However, real-time processing of video stream data in disaster response scenarios entails significant computing costs\cite{koshimura2023digital}, so recent advancements have focused on minimizing computational expense by designing lightweight architectures \cite{piergiovanni2022tiny, tran2019video, 9512294} or selecting representative frames \cite{guo2022deepcore,meng2020ar, bai2024towards}. Building on this inspiration and considering the surplus frame information in raw videos \cite{gao2020listen}, our work aims to create an elegant and efficient model for understanding aerial videos in the context of wildfires.

\begin{figure}[t] 
\centering 
\includegraphics[scale = 0.5]{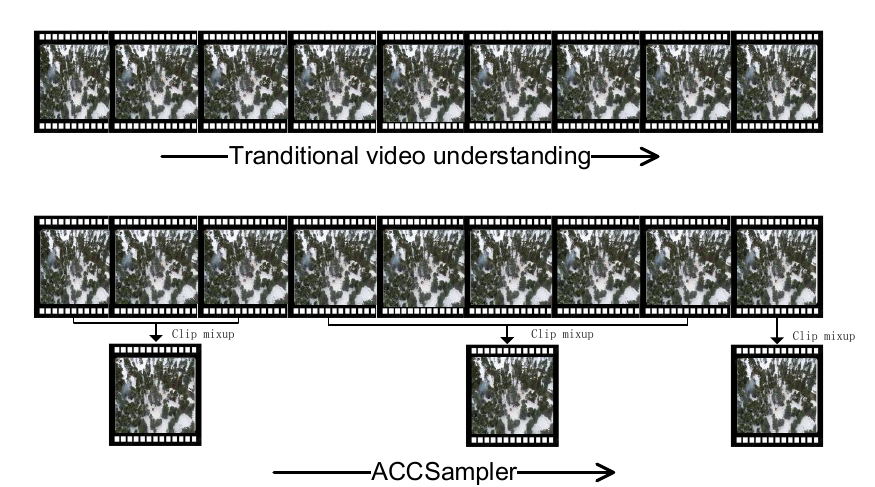}
\caption{Example of ACCSampler for frame consolidation.}\label{overview}
\vspace{-5mm}
\end{figure}

This paper introduces AccSampler, an innovative video understanding model called Adaptive Clip-aware Compression and Frame Sampling Network. Unlike conventional methods that analyze videos frame by frame, AccSampler employs a unique approach by consolidating multiple clips into a single frame, as depicted in Fig.~\ref{overview}. Our model strategically decides when to merge clips, leveraging video features to make informed decisions. We efficiently reduce computational workload without compromising accuracy by dedicating computational resources to frames rich in information and disregarding less relevant clips. AccSampler not only achieves superior accuracy but also substantially minimizes computational efforts.

However, aerial videos are typically captured in challenging environments, where factors such as blurred frames and abrupt changes, like unpredictable camera vibrations due to air currents, are more prevalent. To address this, we introduce the station points to provide future features, aiding the policy network in making sensible and stable decisions when processing aerial videos. The second stage involves the Frame Selection Module. The decisions made by the policy network reflect the frames' importance. We developed a straightforward static frame selection module based on the trained Efficient Video Understanding Module. This module scores each video frame and selects high-scoring frames to reconstruct a more concise video. This process involves no intricate pre-processing or post-processing. It is akin to editing a lengthy video to obtain a shorter, more valuable version. The resulting video preserves essential temporal and spatial information crucial for disaster identification to the greatest extent possible.

The key contributions of this study are outlined below:
\begin{enumerate}

\item{We have designed a new deep-learning model that is both lightweight and efficient, specifically tailored for processing and interpreting aerial videos.}

\item{We have employed the station point approach, providing additional spatio-temporal features. This augmentation enhances the model's efficiency, robustness, and overall performance.}

\item{Through some ablation experiments, we not only confirmed the robustness of our model but demonstrated its applicability for monitoring other physical processes.}

\end{enumerate}

\section{Related Work}

An efficient video understanding model, based on representative frames, offers an effective solution to the challenge of substantial computational costs \cite{lin2022ocsampler}. There are two prominent techniques for frame selection: serial video frame sampling and parallel video frame sampling. In sequential sampling, frames are read in temporal order, and a policy is employed to determine the subsequent frame to be read. AdaFrame \cite{wu2019adaframe} utilizes Memory-augmented LSTM to retain temporal and spatial information and employs hidden layer states to identify the next frame for reading. VideoIQ \cite{9711308} reduces computational overhead by allocating different bits or selecting input resolution for each frame.
On the other hand, parallel sampling processes frames and video clips independently and consolidates the outcomes. SCSampler \cite{korbar2019scsampler} estimates the score of each fixed-length clip and synthesizes predictions. OCSampler \cite{lin2022ocsampler} employs a straightforward single-step reinforcement learning optimization to directly aggregate a more comprehensive set of features for video-level modeling. Furthermore, innovative techniques have emerged in this domain. AdaFuse \cite{meng2021adafuse} dynamically fuses channels from current and past feature maps, enhancing temporal modeling.

The process of frame selection is rooted in the realm of efficient video understanding. For instance, AdaFrame \cite{wu2019adaframe}, Frame Glimpse \cite{yeung2016end}, and SCSampler \cite{korbar2019scsampler} utilize frame sampling techniques coupled with specific policy networks. These approaches discern the most valuable frames and video clips from a sequence, subsequently employing these clips in cut videos for various video recognition tasks. SMART \cite{gowda2021smart} employs a single-frame selector to compute the frame's spatial classification value. It then utilizes a global selector to derive temporal and spatial category values from frame pairs. Ultimately, SMART combines the evaluations of each frame's two values within the video, aiding in frame selection. On the other hand, MGSampler \cite{zhi2021mgsampler} leverages two types of motion representation (motion-sensitive and motion-uniform) to distinguish motion-salient frames from the background efficiently. Using a motion-uniform sampling strategy based on cumulative motion distribution, the MGSampler ensures that the selected frames evenly cover all essential segments with high motion salience.

\section{Proposed Method}
AccSampler aims to achieve efficient video understanding through clip compression, which allows our model to identify insignificant clips and consolidate them into a single frame. Moreover, leveraging the features obtained from clip compression, AccSampler can assess the significance of each frame and select the most informative frames to achieve dataset distillation.

\subsection{Network Architecture}\label{subsec3.1}
\subsubsection{Overview}\label{subsec3.1.1}
\begin{figure*}[t] 
\centering 
\includegraphics[scale = 0.4]{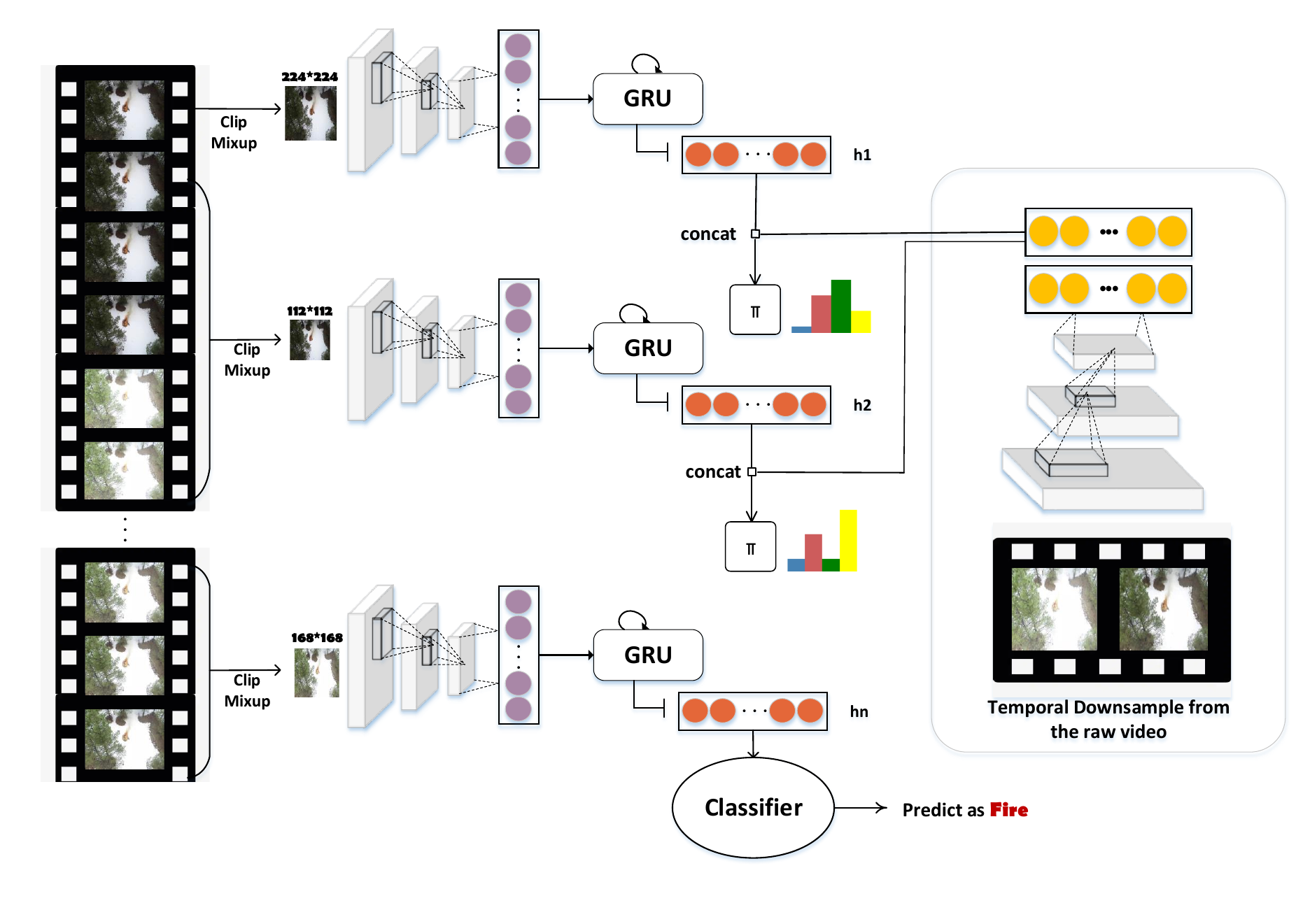}
 \vspace{-3mm}
\caption{The architecture of AccSampler. Frame selection is static and based on the distribution produced by $\pi$}\label{architecture_M1}
\vspace{-5mm}
\end{figure*}

Fig.~\ref{architecture_M1} illustrates the structure of AccSampler. Considering videos $v$ comprising a sequence of frames $X$, denoted as $X$=$\left\{x_t\right\}_{t=1}^{N}$, where N represents the total number of frames in the video, and t denotes the current time step. At t = 1, AccSampler uniformly selects multiple station points from the original input and feeds them into a 2D-CNN to extract station point features, denoted as $\left\{s_{m}\right\}$. These station points provide future information for the policy network $\pi$; at t = i, $\pi$ receives the concatenation of clip-level features from the feature extraction network $f_s$ and the corresponding station point features. Subsequently, $\pi$ evaluates the significance of the following frames and determines how many frames will be compressed at t = i + 1. After processing the entire video, the classifier $f_c$ classifies the video.
Frame selection is static and relies on the well-trained $\pi$. This policy network effectively assesses the importance of subsequent frames. Consequently, based on the output of $\pi$, AccSampler calculates a score for each frame.

\subsubsection{Clip Mixup}\label{subsec3.1.2}
Mixup\cite{zhang2017mixup} is a data augmentation technique that involves scaling two samples and their corresponding labels from the same batch to create a new sample. The mixup formula is as follows:
\begin{align}
    x^{\prime}= \lambda x_i+(1-\lambda)x_j \label{mixup_sample}\\
    y^{\prime}= \lambda y_i+(1-\lambda)y_j \label{mixup_label}
\end{align}
Here, $x_i$ and $x_j$ are raw data inputs, and  $y_i$ and $y_j$ represent their labels. $\lambda$ is the scale parameter, and $\lambda$ follows a $Beta$($\alpha$, $\alpha$) distribution.

We employ clip mixup to consolidate several frames into one to optimize computational efficiency while preserving crucial video information. Since video classification does not demand frame-level labels, clip mixup employs only Eq.~\eqref{mixup_sample}. Specifically, clip mixup merges the initial and final frames of input clips longer than one frame into a single frame $x_i^{mixup}$, representing original clips since frames within a short clip tend to be quite similar.

\subsubsection{Feature Extraction Network}\label{subsec3.1.4}
The feature extraction network $f_s$ combines a lightweight CNN and an RNN. The output of $f_s$ is the hidden state of the RNN. Therefore, the video features up to time t = i can be expressed as:
\begin{equation}
h_i = f_s(x_i^{mixup})\label{hidden state}
\end{equation}
\subsubsection{Policy Network}\label{subsec3.1.5}
The policy network $\pi$ determines the number of frames to be fused at the next time step by selecting action $k_i$ from the discrete action space $A$. Upon receiving the hidden state from $f_s$ at t = i, $\pi$ chooses the closest future station point feature from $\left\{s_m\right\}$. Subsequently, the hidden state and the station point feature are concatenated and input into $\pi$, producing the predicted distribution $a^p_i$ over the action space $A$. This process can be represented as:
\begin{equation}
a^p_i = softmax(\pi(\left[h_i:s_m^{nearest}\right])) \label{p}
\end{equation}
Here, we define $A$ as $\left\{1, 3, 5, 7\right\}$. If $\pi$ makes the decision $k_i$ at t = i, where $k_i\in A$ means that at t = i+1, the $k_i$ frames will be combined into one frame using clip mixup. The smaller the $k_i$, the more crucial the clip is. Additionally, frames of varying importance will be resized to different resolutions. Consequently, we define R=$\left\{224, 168, 112, 84\right\}$, corresponding to the resolution for each action in the action space $A$, which means $\pi$ will determine whether detailed features are needed. If $\pi$ assesses that the current features are sufficient, it may fuse five or seven frames with a lower resolution, while $\pi$ would compress three or one with a higher resolution when detailed features are needed. We designed the structure of $\pi$ as a GroupNorm followed by a single fully connected layer, where GroupNorm is conducive to expediting convergence.

\subsubsection{Frame Selection}\label{frame selection}
$\pi$ is qualified to assess the importance of frames, as reflected in the decisions it makes. The more frames the policy network decides to fuse, the less crucial they are. Consequently, we introduce the preference score S to assess the significance of each frame. According to Eq.~$\eqref{p}$, when $t = i$, the probability distribution of the policy network's prediction for the input in action space A is $p_i = \{a_{ij}^{p}\}$, $1\leq j \leq \|A\|$. Given the clip to be fused at $t = i + 1$ is $clip_{i+1}$, we define the fraction $S_{i+1}$ of $clip_{i+1}$ as:

\begin{equation}
S_{i+1} = \sum^{\|A\|}_{j=1} a_{ij}^{c}/A_j \label{S}
\end{equation}
where c$\in\{p, gm, gs\}$. To refine the preference score S down to the frame level, the score of the middle frame is $S_i$, with a gradual 10$\%$ decay of the score towards the two sides.

\subsubsection{Gumbel Softmax Trick}\label{subsec3.1.8}
Noticeably, the action space $A$ is discrete, making optimizing it with gradient backpropagation challenging. In such cases, reinforcement learning is typically utilized to train the policy network. However, reinforcement learning often suffers from slow convergence in various applications\cite{wu2019liteeval}. To address this issue, we employ the Gumbel Softmax Trick\cite{jang2016categorical}, which optimizes the policy network through gradient backpropagation.
Gumbel Softmax is an effective re-parameterization technique that replaces previously non-differentiable probability distributions with differentiable ones, introducing randomness during training and encouraging the model to explore more. To illustrate, the policy network will determine the action through Gumbel Max (\ref{gumbel max}) instead of an argmax\cite{jang2016categorical}:

\begin{equation}
a^{gm}_i = argmax(\log(a^p_i)+G_i) \label{gumbel max}
\end{equation}
Here, $G_i$=-$\log(-\log(U_i))$, where $U_i\sim Uniform$(0, 1). About backpropagation, a differentiable Gumble-Softmax will be used to approximate argmax to calculate gradients:

\begin{equation}
a_i^{gs} = softmax((\log(a^p_i)+G_i)/\tau) \label{gumbel softmax}
\end{equation}
Here, $\tau$ is a hyperparameter representing the softmax temperature. When $\tau$ $>$ 0, the labels become relatively smooth, while they degrade to one-hot labels when $\tau$= 0. As per prior research\cite{meng2020ar, sun2021dynamic}, we initialize $\tau$ to 5 and anneal it to 0 as the number of epochs increases.

\subsection{Loss Function}\label{subsec3.1.7}
We designed three loss functions to train AccSampler. The first one is a cross-entropy loss, which calculates the classification loss $L_c$ between the prediction and the ground truth:
\begin{equation}
L_c = \mathbb{E}\left[-Y\log(f_c(h_{final}))\right]\label{Lc}
\end{equation}
where $h_{final}$ represents the final hidden states. Additionally, to prevent the policy network from converging to trivial solutions where certain actions are ignored, we introduce the balance loss $L_b$:
\begin{equation}
L_b = \sum_{k\in A}{\mathbb{E}\left[\frac{1}{T}\sum_{t=1}^T\mathbb{I}(k_t=k)-\frac{1}{\|A\|}\right]}\label{Lb}
\end{equation}
where $k_t$$\in$$\left\{1, 3, 5, 7\right\}$ and T represents the total time steps. Lastly, to minimize computation and reach a balance between accuracy and computation, we employ the GFLOPs loss $L_g$, defined as follows:
\begin{equation}
L_g = \frac{1}{T}\sum_{t=1}^T{GFLOPs(f_s(x_i^{mixup}))}\label{Lg}
\end{equation}
Here, we calculate the GFLOPs from the inference of $f_s$, so the $L_g$ of the input V represents the average GFLOPs.
In summary, the total loss $L$ can be defined as:
\begin{equation}
L = L_c+\beta L_b+\gamma L_g \label{loss}
\end{equation}
where $\gamma$ and $\beta$ are hyperparameters that balance the losses. In the experiment, $\beta$ is set to 0.3 and $\gamma$ is set to 0.1 referenced from VideoIQ\cite{9711308}.


\section{Experiments}
\subsection{Experiment Setup}\label{Experiment Setup}
\subsubsection{FLAME Dataset}\label{Dataset}
The FLAME dataset\cite{shamsoshoara2021aerial} is a publicly available collection of fire images and videos captured by drones. We utilized the seventh and eighth repositories of this dataset for image classification. These repositories contain 39,375 images for the training/validation set and 8,617 images for the test set, all with a resolution of $254\times 254$. Notably, we utilized this dataset uniquely for the video classification task. In this context, we defined video samples as sequences of 64 consecutive images extracted from the same video. If any 64 frames were labeled as containing fire, the entire video sample was marked as having fire. 
Consequently, we curated a dataset comprising 615 clips for training and 134 clips for testing. Among these, 483 samples were fire-related, while 266 samples had no fire, maintaining a ratio of approximately 2:1. 



\subsubsection{HMDB51 Dataset}\label{Implementation Details}
The HMDB51 dataset is an open-domain video recognition dataset, encompassing many genuine videos derived from diverse platforms\cite{6126543}. Comprising 6,766 video segments, it covers 51 action categories, ensuring each category has at least 101 clips. This paper uses HMDB51 (split 1) as a supplementary dataset to verify the effectiveness of AccSampler's inefficient video recognition.
\subsubsection{Implementation Details}\label{Implementation Details}
AccSampler sets the number of station points to 2, and the hyperparameter $\alpha$ for clip mixup is 0.3. To optimize computational efficiency, we utilize MobileNet-v2 to construct $f_s$ (the CNN for feature extraction of station points aligns with $f_s$. The GRU's hidden layer dimension is 512, and there is a single layer. Our training approach consists of multiple stages. In the first stage, we employ frame-level labels and fine-tune MobileNet-v2, pre-trained on ImageNet, over 100 epochs. The initial learning rate is set to 0.01 and is reduced by a factor of 10 at epochs 50, 70, and 90. In the second stage, we freeze MobileNet-v2 and train GRU using clip-level labels for 20 epochs, starting with an initial learning rate of 1.45e-5. In the third stage, MobileNet-v2 and GRU remain frozen, and we train the policy network with an initial learning rate of 0.01.
We employ $a_i^{gm}$ to calculate the preference score for frame selection. We then train ResNet18-TSM on the selected frames to evaluate the frame selection's performance. We train ResNet18-TSM for 10 epochs, initiating with a learning rate of 0.001 and a weight decay 3e-6.

\subsubsection{Evaluation Metrics}\label{Metrics}
We consistently employ accuracy as the metric for the video classification task. Additionally, computational cost is measured in giga floating-point operations per second (GFLOPs), which can be used to calculate the complexity of the model.

\subsection{Main Result}\label{main result}
\subsubsection{Video Classification}\label{video classification}
We use $f_s$ and $\pi$ to perform video classification, comparing them with GRU(i.e., $f_s$ only). The results are outlined in Table $\ref{performace of cla}$.

\begin{table}[ht]
\caption{Video classification results on the FLAME dataset.}\label{performace of cla}
\vspace{-2mm}
\centering
\begin{threeparttable}[b]
\begin{tabular}{ccccc}
\toprule
Model & Test acc  & Training acc & GFLOPs/v\tnote{1} & GFLOPs/f\tnote{2}\\
\hline
GRU & 79.10 &\textbf{ 98.61} & 26.28 & 0.411\\
\textbf{GRU+policy} & \textbf{82.84} & 97.40 & \textbf{2.79} & \textbf{0.044}\\ 
\bottomrule
\end{tabular}
\begin{tablenotes}
\item[1]{GFLOPs/v refers to the average GFLOPs on the video level}
\item[2]{GFLOPS/f refers to the average GFLOPs on the frame level} 
\end{tablenotes}
\end{threeparttable}
\end{table}

Notably, our model substantially enhances test accuracy. At the same time, the computational load (measured in GFLOPs) is approximately one-tenth that of GRU, meaning the policy network learns to pay closer attention to informative frames.

\subsubsection{Frame Selection}\label{Frame Selection}
The distilled dataset is used to train a sophisticated classification model TSM\cite{lin2019tsm} to evaluate the performance of frame selection. The results are given in Table $\ref{acc on TSM}$.

\begin{table}[ht]
\caption{Test accuracy($\%$) of TSM trained on the different number of frames selected by four methods on FLAME dataset}\label{acc on TSM}%
\vspace{-2mm}
\centering
\begin{threeparttable}[b]
\begin{tabular}{ccccc}
\toprule
 & Uniform\cite{lin2022ocsampler}  & Random\cite{lin2022ocsampler} & SMART\tnote{1}\cite{gowda2021smart} & AccSampler\\
\midrule
8 & 87.3 & 85.1 & 84.3 & \textbf{87.3}\\
20 & 88.1 & 88.1 & 88.1 & \textbf{91.0}\\
28 & 84.3 & 86.6 & 88.8 & \textbf{91.0}\\
\midrule
All(64) & \multicolumn{4}{c}{88.1}\\
\bottomrule
\end{tabular}
\begin{tablenotes}
\item[1]{Our implementation because of no official code}
\end{tablenotes}
\end{threeparttable}
\end{table}

It can be seen that AccSampler demonstrates superior effectiveness, maintaining classification accuracy even when using just 8 frames, surpassing the results obtained from analyzing the full video. AccSampler significantly outperforms the baseline (Uniform and Random) and SMART results when considering an equivalent number of frames. This outcome implies that AccSampler adeptly selects more representative frames, facilitating a more efficient and accurate classification of fire videos.

\begin{figure}[t]%
\centering
\vspace{-5mm}
\includegraphics[scale = 0.45]{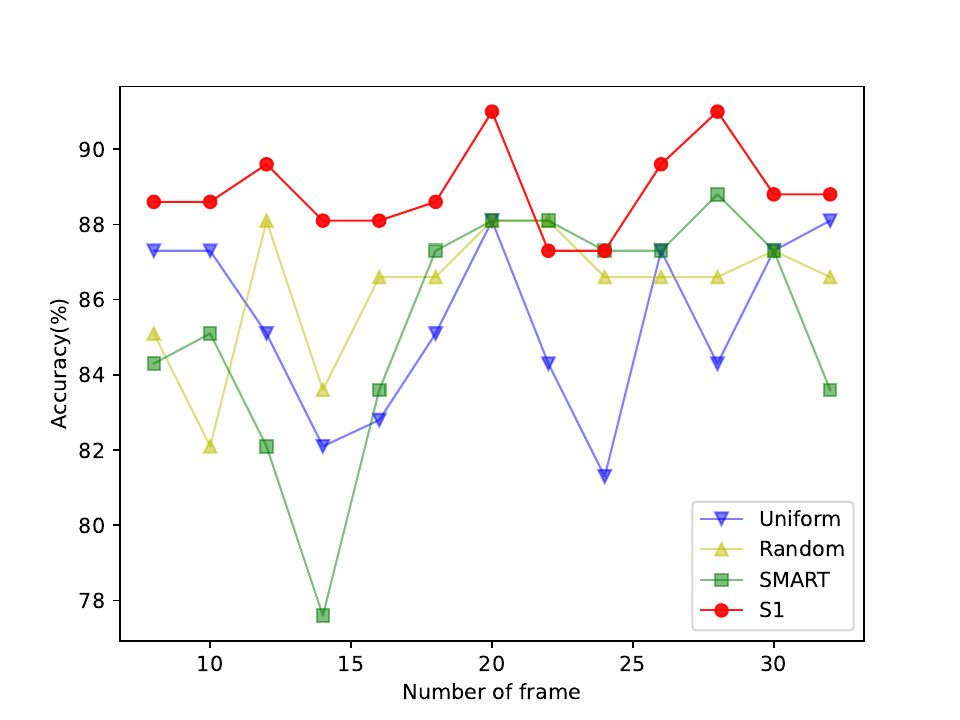}
\vspace{-3mm}
\caption{Comparison of frame selection methods at various frame rates.}\label{fig2}
\vspace{-3mm}
\end{figure}

Further tests were conducted over a broader range of frames to investigate whether the observed results were influenced by data randomness due to the dataset's limited size, as shown in Fig.~\ref{fig2}. The findings reveal that AccSampler consistently outperforms the other three frame selection methods.

However, it's noteworthy that SMART did not exhibit a significant performance advantage over other models. This lack of distinction might be attributed to FLAME's small size and homogeneity, causing complex models like SMART to suffer severe overfitting during training.

\subsubsection{Classification Results on the HMDB51 Dataset}\label{Domain Adaption}
To verify the generalization of AccSampler, we further extend the experiment in Table. \ref{performace of cla} on the HMDB51 dataset. The results are shown in the Table. \ref{performace of cla in hmdb}.

\begin{table}[ht]
\caption{Video classification results on the HMDB51 dataset}\label{performace of cla in hmdb}
\vspace{-2mm}
\centering
\begin{threeparttable}[b]
\begin{tabular}{ccccc}
\toprule
Model & Test acc  & Training acc & GFLOPs/v & GFLOPs/f\\
\hline
GRU & \textbf{40.72} &\textbf{95.78} & 30.19 & 0.32\\
\textbf{GRU+policy} & 38.50 & 92.34 & \textbf{1.29} & \textbf{0.013}\\ 
\bottomrule
\end{tabular}
\end{threeparttable}
\end{table}

Although accuracy declines modestly (about 2


\section{Discussion}
\subsubsection{Preference Score S}
\begin{figure}[h]%
\centering
\vspace{-3mm}
\includegraphics[scale=0.5]{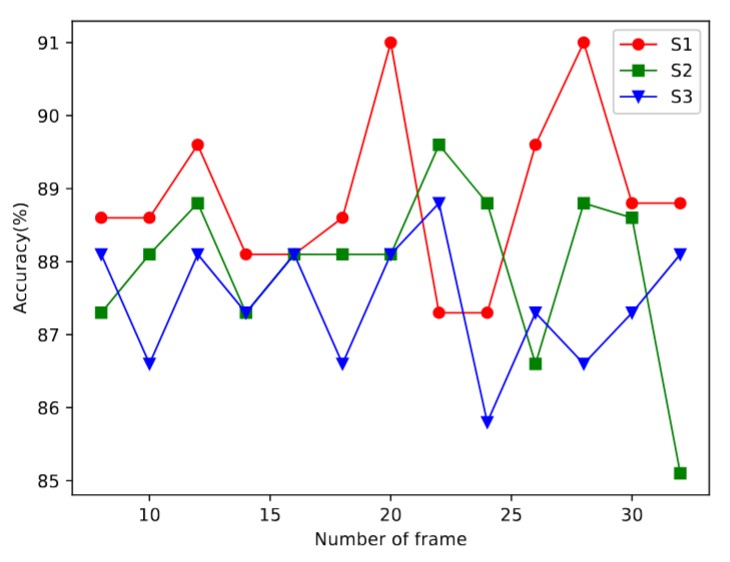}
\vspace{-3mm}
\caption{Comparison of different calculation methods for S.}\label{fig3}
\end{figure}

Apart from the calculation utilizing  $a_i^{gm}$ in Eq.~$\eqref{S}$, we also explored the computation of S using $a_i^p$ in Eq.~$\eqref{p}$ and $a_i^{gs}$ in Eq.~$\eqref{gumbel softmax}$. The scores derived from these three methods are denoted as S1, S2, and S3, respectively. We conducted experiments to evaluate the performance of these three approaches, and the results are depicted in Fig.~\ref{fig3}. The results indicate that the S1 score exhibits significantly superior performance.

\subsubsection{Effect of the Action Space}\label{Effect of the Action Space}
We investigated the impact of the action space $A$ on the classification performance, and the results are shown in Table \ref{effect of action space}. Comparative analysis reveals that the model achieves the highest test set accuracy and the lowest computational load when the action space is defined as $\left\{1, 3, 5, 7\right\}$. Notably, when extending the action space to $\left\{1, 3, 5, 7, 9\right\}$, AccSampler sees an increase in GFLOPs and a drop in the test accuracy, which means broader action space makes the policy network more conservative instead. 

\begin{table}[h]
\caption{The effect of different action spaces on video classification}\label{effect of action space}
\vspace{-2mm}
\centering
\begin{threeparttable}[b]
\begin{tabular}{ccccc}
\toprule
$A$ & Test acc  & Training acc & GFLOPs/v & GFLOPs/f\\
\midrule
$\left\{1, 3, 5\right\}$\tnote{1} &\textbf{82.84} &97.24 & 3.41 & 0.053\\
$\left\{1, 3, 5, 7, 9\right\}$\tnote{2} & 77.60 & \textbf{97.56} & 3.26 & 0.051\\
$\left\{\textbf{1, 3, 5, 7}\right\}$ & \textbf{82.84} & 97.40 & \textbf{2.79} & \textbf{0.044}\\
\bottomrule
\end{tabular}
\begin{tablenotes}
\item[1]{Here R=$\left\{224,168,112\right\}$}
\item[2]{Here R=$\left\{224,168,140,112,84\right\}$}
\end{tablenotes}
\end{threeparttable}
\end{table}


\subsubsection{Effect of Hyperparameter in Clip Mixup}
Variations in the parameter $\alpha$ in clip mixup impact the scale parameter $\lambda$, influencing the model's performance. The findings in Table \ref{effect of alpha} indicate that the model exhibits relatively low sensitivity to changes in $\alpha$. Optimal performance across all metrics is observed when $\alpha$ is set to 0.3.

\begin{table}[h]
\caption{Effect of different $\alpha$ on video classification}\label{effect of alpha}
\vspace{-2mm}
\centering
\begin{tabular}{ccccc}
\toprule
 $\alpha$ & Test acc  & Training acc & GFLOPs/v & GFLOPs/f\\
\midrule
0.2 & 80.60 &97.07 & 2.827 & 0.0442\\
0.4 &\textbf{ 82.84} & 97.07 & 2.831 & 0.040\\
0.5 &81.36 & 97.24 & 2.811 & 0.0439\\
\textbf{0.3} & \textbf{82.84} &\textbf{ 97.40} &\textbf{ 2.792} & \textbf{0.044}\\ 
\bottomrule
\end{tabular}
\end{table}

\subsubsection{Effect of the Station Points}
We experimented with varying the number of station points to assess their impact on classification, and the results are detailed in Table \ref{effect of station points}. The outcomes of the policy network with and without station points exhibit notable differences. Introducing just one station point leads to a 5$\%$ increase in test accuracy and a significant reduction in GFLOPs.
In contrast, the policy network without station points tends to adopt a more conservative approach. As the number of station points increases, the policy network is inclined to fuse more frames. Ultimately, the policy network strikes a balanced trade-off between computation and accuracy with two station points. 

\begin{table}[h]
\caption{Effect of different number of station points on video classification}\label{effect of station points}
\vspace{-2mm}
\centering
\begin{tabular}{ccccc}
\toprule
  & Test acc  & Training acc & GFLOPs/v & GFLOPs/f\\
\midrule
0 & 75.37 &\textbf{98.2} & 6.61 & 0.103\\
1 & 80.60 & 98.1 & \textbf{2.54} & \textbf{0.040}\\
\textbf{2} & \textbf{82.84} & 97.40 & 2.79 & 0.044\\ 
3 &78.36 & 97.7 & 3.06 & 0.048\\
\bottomrule
\end{tabular}
\end{table}




\section{Conclusion}\label{Conclusion}
In this study, we introduced AccSampler, a novel approach to enhancing the efficiency of aerial video understanding in wildfire monitoring tasks. Extensive experiments have demonstrated that AccSampler outperforms several baseline methods in fire detection, offering an excellent distilled dataset for TSM to achieve high precision with low computational costs. More importantly, AccSampler's compression mechanism is model-agnostic and compatible with various backbones, meeting the standards of diverse video-related disaster response tasks.

\bibliography{manuscript}
\bibliographystyle{IEEEtran}

\end{document}